%% file: main.tex
\documentclass[runningheads]{llncs}

\usepackage{graphicx}
\usepackage{amsmath}
\usepackage{amssymb}
\usepackage{booktabs}
\usepackage{multirow}
\usepackage{enumitem}
\usepackage{subfloat}
\usepackage{tikz}
\usepackage{comment}
\usepackage{amsmath,amssymb} %
\usepackage{color}

\usepackage[accsupp]{axessibility}  %
\usepackage[pagebackref,breaklinks,colorlinks]{hyperref}

\input{latex/macros}

\newcommand{\height}{{Pixel Height}}

\begin{document}
\pagestyle{headings}
\mainmatter
\def\ECCVSubNumber{5932}  %

\title{Controllable Shadow Generation\\Using \height~Maps\thanks{Y. Sheng and Y. Liu contributed equally.}} %

\titlerunning{}
\author{Yichen Sheng$^*$\inst{1} \and
Yifan Liu$^*$\inst{2} \and
Jianming Zhang \inst{3} \and 
Wei Yin \inst{2} \and 
A. Cengiz Oztireli \inst{4} \and
He Zhang \inst{3} \and 
Zhe Lin \inst{3} \and 
Eli Shechtman \inst{3} \and
Bedrich Benes \inst{1} 
}
\authorrunning{Y. Sheng et al.}
\institute{Purdue University \and 
University of Adelaide \\ \and 
Adobe Research \and  
University of Cambridge}

\maketitle

\begin{abstract}
\input{paragraph/abstract}
\end{abstract}

\section{Introduction}
\label{sec:intro}
\input{paragraph/Introduction}
\section{Related Work}
\label{sec:rel}
\input{paragraph/related_work}

\section{Method}
\input{paragraph/method}

\section{Experiments and Evaluation}
\label{sec:med}

\input{paragraph/experiment}

\section{Conclusion}
We proposed an approach for generating controllable perceptually plausible shadows based on the \height~map. The new geometry representation, \height~map, encodes the correlations among objects shape, camera pose, and the ground. It can be directly labeled or estimated from 2D images. The position and softness is controlled in an easy interactive way. Qualitative and quantitative comparisons demonstrate the results and generalization ability of the proposed method outperforms previous deep learning-based shadow generation methods. However, our \height~map representation only considers the frontal surface of the object. This assumption works when the shadow of the object's back surface mostly overlaps with that of the frontal surface. A thickness map defined on each point may address this issue and is worth future exploration. 

\noindent\textbf{Acknowledgment} Most of the work was done during Yifan and Yichen's internship at Adobe. This work was also supported by a UKRI Future Leaders Fellowship [grant number G104084]. We thank Dr. Zhi Tian for the discussions.

\clearpage
{\small
\bibliographystyle{splncs04}
\bibliography{egbib}
}
\end{document}

%% file: latex/macros.tex
\usepackage{color}
\usepackage{multirow}
\usepackage{xcolor}

\newlength\savewidth

\definecolor{turquoise}{cmyk}{0.65,0,0.1,0.1}
\definecolor{purple}{rgb}{0.65,0,0.65}
\definecolor{darkgreen}{rgb}{0.0, 0.5, 0.0}
\definecolor{darkred}{rgb}{0.5, 0.0, 0.0}
\definecolor{darkblue}{rgb}{0.0, 0.0, 0.5}
\definecolor{blue}{rgb}{0.0, 0.0, 1.0}
\definecolor{orange}{rgb}{1.0,0.5,0.0}

\newcommand{\hide}[1]{{}}

\makeatletter
\renewcommand{\paragraph}{%
  \@startsection{paragraph}{4}%
  {\z@}{0.3ex \@plus 1ex \@minus .1ex}{-1em}%
  {\normalfont\normalsize\bfseries}%
}
\makeatother

\newif\ifproofread

\newcommand{\newcontent}[1]{%
\ifproofread
\textcolor{blue}{#1}%
\else
#1%
\fi
}

%% file: paragraph/abstract.tex
Shadows are essential for realistic image compositing from 2D image cutouts. Physics-based shadow rendering methods require 3D geometries, which are not always available. Deep learning-based shadow synthesis methods learn a mapping from the light information to an object's shadow without explicitly modeling the shadow geometry. Still, they lack control and are prone to visual artifacts. We introduce ``\height", a novel geometry representation that encodes the correlations between objects, ground, and camera pose. The \height~can be calculated from 3D geometries, manually annotated on 2D images, and can also be predicted from a single-view RGB image by a supervised approach. It can be used to calculate hard shadows in a 2D image based on the projective geometry, providing precise control of the shadows' direction and shape. Furthermore, we propose a data-driven soft shadow generator to apply softness to a hard shadow based on a softness input parameter. Qualitative and quantitative evaluations demonstrate that the proposed \height~significantly improves the quality of the shadow generation while allowing for controllability.

%% file: paragraph/Introduction.tex
\begin{figure}[t]
    \centering
    \includegraphics[width=1\textwidth,height=3cm]{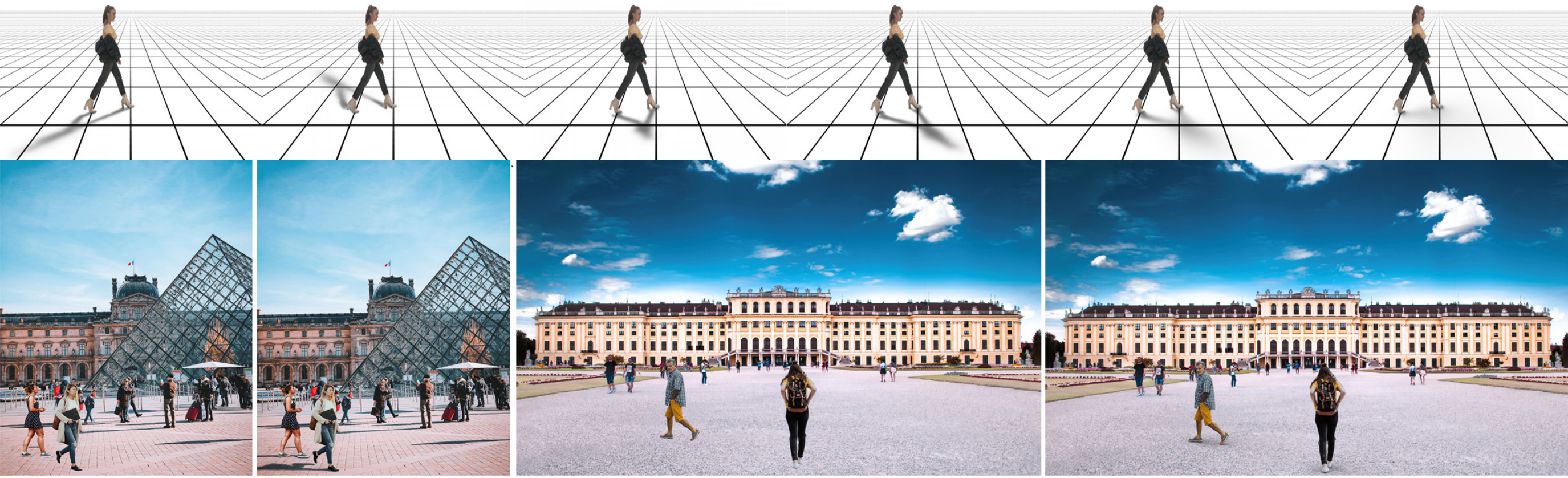}
    \caption{Controllable shadow generation with the proposed method. \textbf{First row}: With the help of our new introduced \textit{Pixel Height} of an object, users can control the position and the softness of the generated shadows. \textbf{Second row}: The composited images with our generated shadows (2nd and 4th) are much more natural than the direct composites (1st and 3rd).}
    \label{Fig: first page fig.}
\end{figure}

Shadow generation is an important step for image composting that enhances photo realism and adds positional and directional cues for the composed objects. Advanced image editing techniques enable composing objects into a new background with accurate segmentation and matting~\cite{lu2019indices} and harmonization of color styles~\cite{jiang2021ssh}. However, the composited objects are not realistic if no matching shadows are synthesized (see the 1st and 3rd images in the second row in Fig.~\ref{Fig: first page fig.}). Manually creating a perceptually plausible shadow for a 2D object is tedious, even for an experienced artist, especially for extended (linear or area) light sources. 

Mature techniques that calculate soft shadows for 3D scenes exist~\cite{crow1977shadow,oztireli2016integration}. However, 3D shape information is often unavailable when we composite objects from real images. Recent deep learning advancements brought significant progress to shadow generation in 2D images. A series of methods~\cite{hong2021shadow,liu2020arshadowgan,zhang2019shadowgan} based on generative adversarial networks (GANs) have been proposed to automatically generate shadows by training with pairs of shadow and shadow-free images. 
These methods mainly focus on generating hard shadows, and the final results are not editable. \newcontent{Moreover, these methods require the background scene to implicitly provide light information, while in many application scenarios, objects are either composited on abstract or pure color background. Also, shadow editing needs to be applied on separate image layers with background images missing or incomplete at the time of editing. Therefore, shadow generation for object cutouts with user control is more suited for professional image editing workflows.}
Recently, Sheng et al.~\cite{sheng2021ssn} proposed to learn a mapping from a 2D  cutout of the object to the corresponding soft shadows based on a controllable lightmap and achieved promising results. However, \newcontent{due to the lack of geometry guidance}, this method cannot generalize well for varying scenes and may lead to visible artifacts in the generated shadows.

\begin{figure}[hbt]
\centering  %
\includegraphics[width=\linewidth]{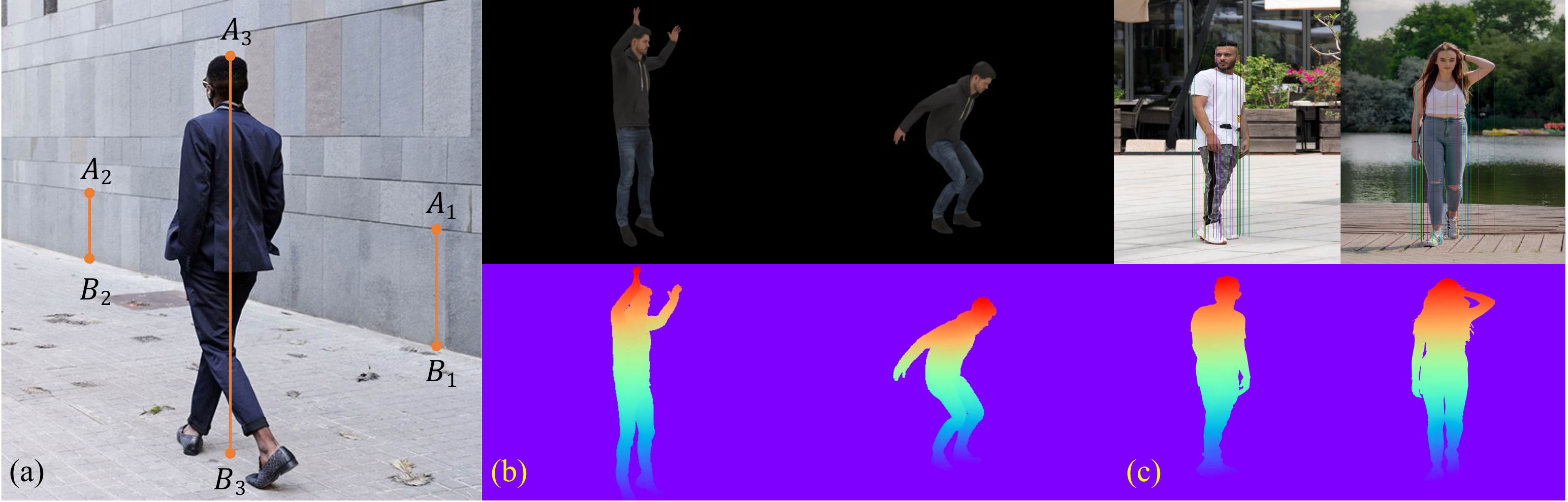}
\caption{\textbf{\height.} (a). The number of pixels between point $A_i$ and $B_i$ is the \height~for point $A_i$. We collected two datasets with \height~annotation: Synthetic60K and Real1500. (b) shows the sample data with various poses from 3D models. (c) shows samples of the sparsely labeled data and the interpolated \height~map. Note that every \height~map is divided by its max-height value for visualization.} 
\label{Fig.real}
\end{figure}

\newcontent{We introduce a controllable and editable shadow generation method for 2D object cutouts.} We introduce \emph{\height}, a new 2.5D shape representation for an image to provide geometry guidance.  
The \height~is defined as the pixel distance between a point on an object and its \emph{footpoint}, namely its vertical projection on the ground in the image (see Fig.~\ref{Fig.real}-(a)). Based on \height, we can explicitly compute the shadow point based on projective geometry. The \height~could be measured and annotated on a 2D image or calculated from synthetic data with 3D object models. Similar to monocular depth estimation, \height~can also be estimated from a single RGB image by a data-driven method. We \newcontent{collect synthetic and real annotated data (see Fig.~\ref{Fig.real}-(b) and (c)) to} train a \height~map prediction model for object cutouts.

Given the annotated or predicted \height~map of an object, we render a hard shadow based on the position of the horizon and the point light in the \newcontent{2D} image space with a proposed \textit{hard shadow renderer}. \newcontent{To add softness to the shadow,} we learn \newcontent{an efficient and controllable} mapping from the hard shadow to the soft shadows based on a softness parameter using \textit{a soft shadow generator}. As shown in Fig.~\ref{Fig: first page fig.}, our system can generate varying shadow maps controlled by the light source position and the softness control. Our method explicitly models the shadow geometry that is more controllable and robust than methods that directly predict shadows based on an image background or a light map. 

We conduct extensive experiments to show that the \height~map improves the controllability of shadow generation. Realistic shadows are synthesized by easy and intuitive user control given the RGB image and an object segmentation mask. Qualitative and quantitative results demonstrate that our method generates higher quality shadows than previous interactive and automatic shadow generation algorithms in 2D images. Our main contributions are:
\begin{itemize}[noitemsep]
    \item A formulation of hard shadow rendering in images based on a novel geometry representation, \height, which can be manually labeled or predicted by a model from a single image.
    \item A controllable shadow generation framework, where users control the position and softness of an object shadow. The framework consists of a \height~estimation, a hard shadow renderer and a soft shadow generator.
    \item Extensive evaluation and analysis, showing improved quality and controllability of our proposed \height~based shadow synthesis method.
\end{itemize}

%% file: paragraph/related_work.tex
\noindent\textbf{Shadow rendering in graphics.} Shadow rendering based on 3D geometries is a well-studied technique in computer graphics. In real-time rendering, shadow volume~\cite{assarsson2003geometry,schwarz2007bitmask} and shadow map based rendering techniques are mainstream approaches~\cite{donnelly2006variance,reeves1987rendering,sen2003shadow,williams1978casting}. The soft shadow is approximated either by blurring the hard shadow boundaries~\cite{annen2008real,chan2003rendering,fernando2005percentage,guennebaud2006real,guennebaud2007high,soler1998fast} or weighted sum of a set of hard shadows sampling on an area or volume light source~\cite{crow1977shadow}. Many works~\cite{franke2014delta,mehta2012axis,oztireli2016integration} have been proposed to speed up this sampling process by adjusting the density and the weight. Besides, some simplified geometries~\cite{fuchs1985fast,ren2006real} or light representations~\cite{heitz2016real} are proposed to render shadows in real-time. Global illumination algorithms \cite{cook1984distributed,kajiya1986rendering,ng2003all,sillion1991global,westin1992predicting} render soft shadows implicitly. Such methods can render realistic shadows for complicated objects given accurate 3D object models. However, 3D models are not always available for objects in real images, especially in image compositing tasks in computer vision. 

\noindent\textbf{Shadow synthesis with deep learning.} In recent years, generative adversarial networks (GANs) have achieved significant improvements on image translation tasks~\cite{isola2017image,liu2018auto}. A series of works~\cite{hong2021shadow,hu2019mask,liu2020arshadowgan,zhang2019shadowgan} have been introduced for generating shadows directly from a composited shadow-free image based on the object mask guidance. ARShadowGan~\cite{liu2020arshadowgan} renders a dataset by inserting 3D objects into real background images with augmented reality. Hong~\textit{et al.}~\cite{hong2021shadow} generate the shadow-free images by removing the shadow region from the real-world images. These methods try to predict the style and the color of the final shadow by a data-driven method, but they cannot provide controllability for the user. %

Sheng~\textit{et al.}~\cite{sheng2021ssn} propose an interactive soft shadow generation network based on a user-provided lightmap. Physics-based methods on 3D object models render their training data. The network is trained to learn the mapping from the 2D object cutout and environment lightmap to the soft shadow maps. Contrary to the previous works, we generate soft shadows by first generating a hard shadow and converting it to a soft shadow using a softness input parameter. This hard-to-soft transformation is much easier to learn. The hard shadow can be obtained with our proposed pixel-height map by calculating the occlusion directly in the 2D projection space with a simple shadow projection model. 

\noindent\textbf{Geometry representation.}
Similar to monocular depth estimation~\cite{saxena2005learning}, recovering \height~map from a single image is an ill-posed problem. Numerous methods~\cite{li2018megadepth,yin2021virtual,yin2019enforcing} exist to estimate depth from a single view image by supervised methods. As the depth is a 2.5D representation, the intrinsic camera parameters are required for recovering the 3D shape of the object. The 3D point cloud~\cite{remondino2003point} is another geometry representation for 3D objects' shape. They can be captured by special scanners, recovered from multi-view images, but cannot be labeled directly just from a monocular image. 
Furthermore, methods~\cite{pifuSHNMKL19,saito2020pifuhd} have been proposed to directly recover the 3D shape, especially for humans from a monocular image. The proposed \height~map is a new geometry representation, which reflects the correlation among the object, shadow receiver, and camera pose. It is easier to interpret and annotate, and it is useful for applications that require explicit occluder-receiver constraints such as shadow generation.

%% file: paragraph/method.tex
\label{sec:med}

We propose a new approach to generate perceptually plausible soft shadows on 2D images without 3D object models. The key idea of our approach is to render the object's hard shadow from a point light in the image plane following a simplified projective geometry constraint  (see Sec.~\ref{sec:hard_render}), and then synthesize the corresponding soft shadow based on the hard shadow using a data-driven approach (see Sec.~\ref{sec:soft_render}). 
\begin{figure}[hbt]
\setlength{\belowcaptionskip}{-0.2cm}
\centering  %
\includegraphics[width=1\linewidth]{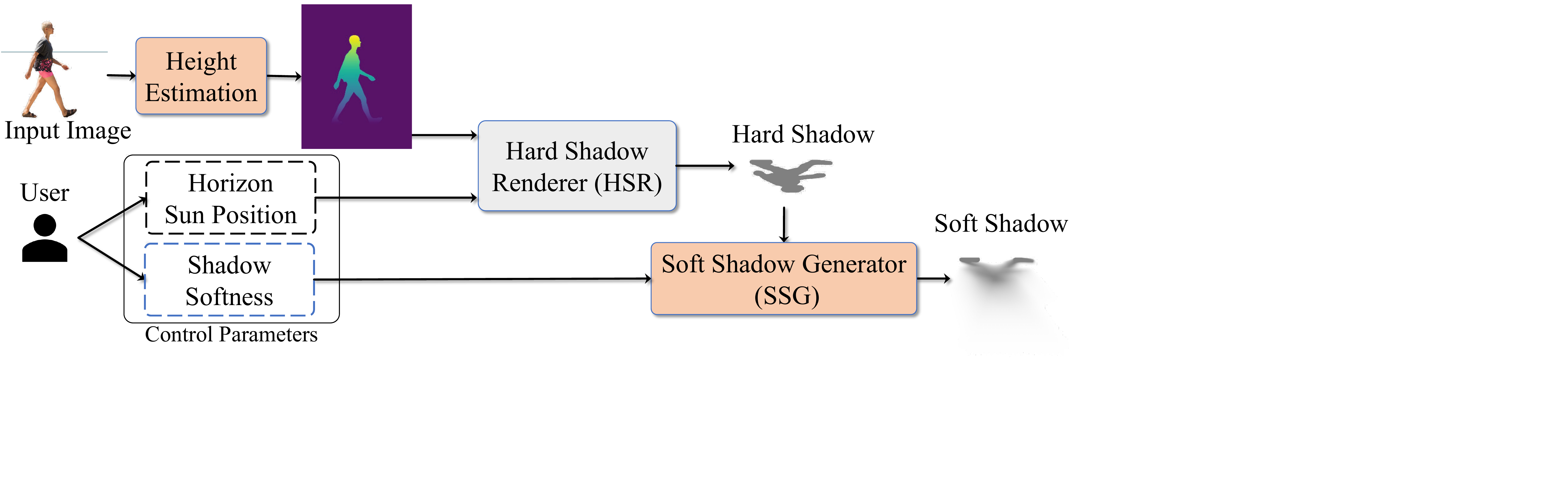}
\caption{
Given a 2D foreground image and its \height~map, a user can control the position and the softness of the generated shadow by the user-defined light information or the existing image-based light information. The \height~map can be manually annotated on images or predicted from a single image by training a model.}
\label{fig:pipeline}
\end{figure}

We need to know the shape of the object and its relationship with the shadow receiver and the camera to render a hard shadow on the image plane. A~new geometry representation \height~is proposed to represent the object shape in 2D images, which is essential to rendering the hard shadow. We verify that this representation can be estimated by a data-driven approach (see Sec.~\ref{sec:height_prediction}). 

As shown in Fig.~\ref{fig:pipeline}, given the foreground object image and its mask, we can annotate or estimate its \height~map. The hard shadow's position and shape can be determined by the controllable light information (the sun position) and the ground (the horizon line). Finally, based on the hard shadow, the soft shadow 
generator can produce a perceptually pleasing soft shadow according to the softness control parameter. The user could provide all the controllable variables with a simple GUI (see the supplementary videos), but they can also be potentially estimated from the background image.

\subsection{Hard Shadow Renderer in 2D Image}\label{sec:hard_render}

This section introduces our novel hard shadow rendering method based on the following assumptions: (1) images are upright, and the vertical lines are parallel. This corresponds to the one-point perspective or the two-point perspective, which is very common, and (2) the light source is a point light and is always above ground. \newcontent{Note that if the first assumption does not exactly hold, the generated hard shadow will be slightly distorted, but still a good approximation.}

\begin{figure}[t]
\centering  %
\includegraphics[width=0.85\linewidth]{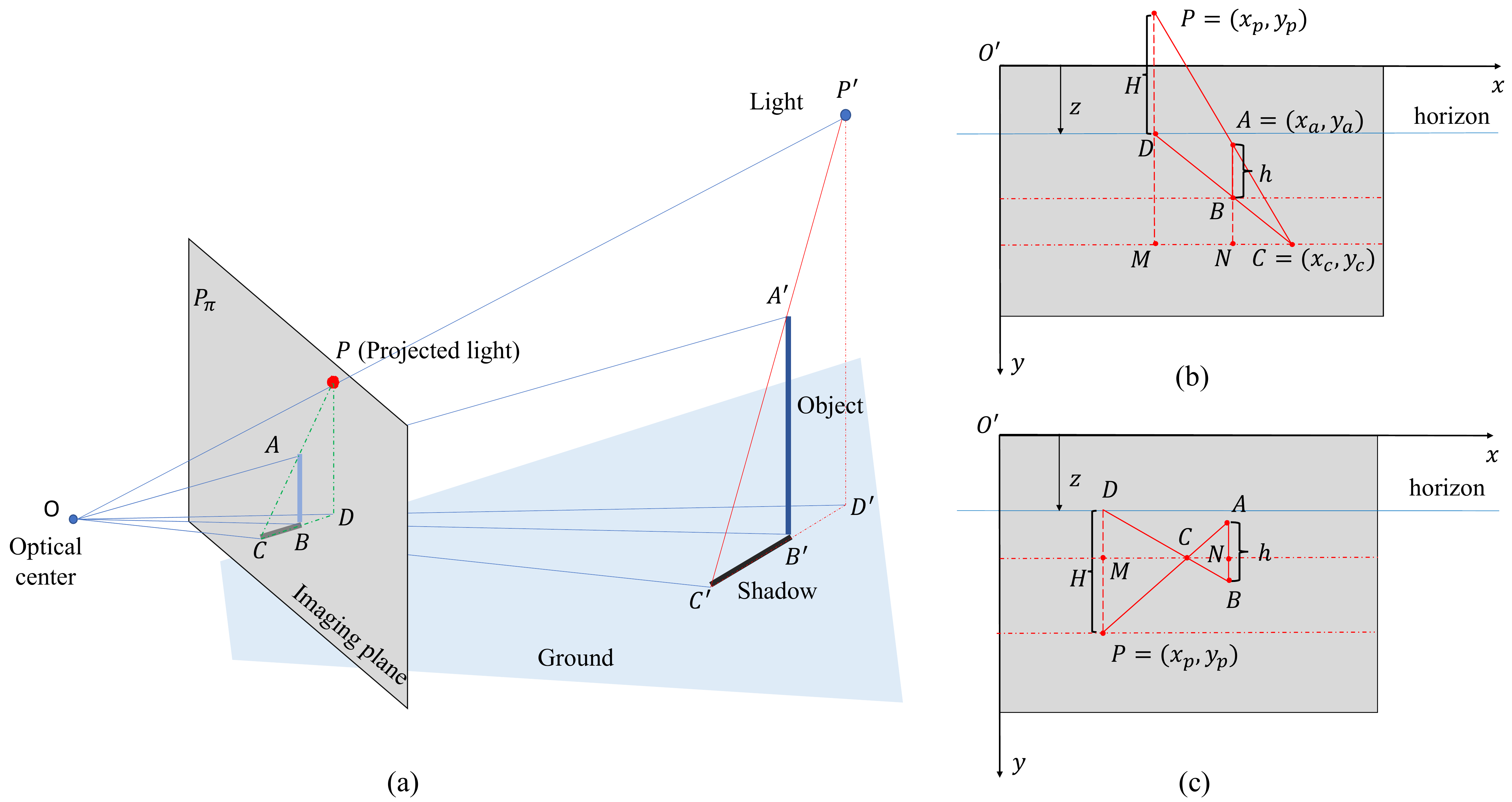}

\caption{\newcontent{\textbf{Hard shadow renderer using \height~representation.} (a) shows the camera model, where $A'B'$ is the object standing upright on the ground, $P'$ is the point light source and $C'$ is the shadow point of $A'$. (b) and (c) shows two typical cases of the projection of the light, the object, and its shadow on the image plane. 
$D$ is the perpendicular feet of $P$. The intersection point $C$ of $DB$ and $PA$ is a projection of the shadow point $C'$. The 3D collinear property in the shadow geometry is preserved after being projected to the image plane. } }

\label{Fig.hard shadow renderer}

\vspace{-1em}
\end{figure}

\newcontent{A simple example of the projective geometry following our assumptions is shown in Fig.~\ref{Fig.hard shadow renderer}. Given an object $A'B'$ that stands vertically on the ground and a point light source $P'$, the object's shadow is then cast to~$B'C'$. Given an image plane, the light source, the object, and the shadow are projected to $P$, $AB$, and $BC$, respectively. The point $D'$ is the perpendicular footpoint of the light, which is projected to $D$. Note that $P'$, $A'$ and $C'$ are always collinear; and $C'$, $B'$ and $D'$ are always co-linear. Thus, the projections of these points are also collinear in the image plane. 
For a non-planar shadow receiver, \emph{e.g.}~a wall, a similar collinear condition still holds except that the shadow point $C'$ will be above the ground, and it will have its footpoint. For simplicity, we study the special case where the shadow receiver is the ground plane in the following. A more general formulation can be found in the supplementary material.
}

Fig.~\ref{Fig.hard shadow renderer} (b) and (c) show the image plane and the relevant variables. We define the upper left corner of the image as the origin of the coordinate system.
The light $P$ and its projected perpendicular footpoint $D$ are located at $(x_{p}, y_{p})$ and  $(x_{p}, y_{p}+H)$ respectively, where $H$ is the pixel distance between $P$ and its footpoint, and we call it the \emph{\height}~of the light.
Similarly, the object point $A$, its footpoint $B$ and its shadow point $C$ are located at $(x_{a}, y_{a})$, $(x_{a}, y_{a}+h)$ and $(x_c, y_c)$, where $h$ is the \height~of $A$. 

According to the triangle similarity in Fig.~\ref{Fig.hard shadow renderer}-(b) and Fig.~\ref{Fig.hard shadow renderer}-(c), we have
\begin{equation}
    \frac{h}{H} = \frac{CN}{CM} = \frac{x_c-x_a}{x_c-x_p} 
                = \frac{AN}{PM} = \frac{y_c-y_a}{y_c-y_p}.
\end{equation}
The shadow point $C$ can be derived from $(x_a, y_a, h)$ and $(x_p, y_p, H)$ by
\begin{equation}
    C=[x_{c},y_c]= \frac{1}{H-h}\left[Hx_a-hx_p,Hy_a-hy_p\right].
\label{eq:shadow}
\end{equation}

Note that $H$ may take a positive or a negative value. A negative value of $H$ indicates that the light is behind the camera, and the shadow will be cast away from the camera (see Fig.~\ref{Fig.hard shadow renderer}-(c)). Note that the derived $C$ may not exist. For example, when $h>H>0$, the ray $\overrightarrow{PA}$ will not intersect with the ground. In this case, the derived $C$ is actually the ground intersection point in the opposite direction of the ray.  
There is a special case when the light is infinity, and its footpoint is on the horizon. Let $Z$ denote the y coordinate of the horizon. In this case, we can replace $H$ with $Z-y_p$ in Eq.~\ref{eq:shadow}, and control the perspective of the shadow using the horizon line (see Fig.~\ref{Fig.hard}).

\newcontent{The above formulation describes how the shadow geometry is derived for our \height~representation. For generic scenarios, the \height~map of the shadow receiver needs to be provided, and the shadow map can be calculated by checking the collinear conditions similar to the ones mentioned above. We implemented the rendering algorithm for generic shadow receivers using CUDA. Please refer to supplementary materials for details. The visibility of the pixels on the shadow receiver can be computed in 20 ms for a $512 \times 512$ image. }

\subsection{Pixel Height Map Estimation} \label{sec:height_prediction}

A \height~map is a 2.5D representation similar to a depth map. Different from the depth map, the \height~map uses the ground plane as a world frame reference to locate the object. It captures the object-ground relation so that the contact points and the uprightness of the object are explicitly enforced. In addition, \height~map is measurable in the image space and can be annotated manually. 
In contrast, traditional 2.5D representations like depth are challenging to annotate from a single image. Objects reconstructed from a depth map can also be tilted if the camera intrinsic parameters are unknown. 

The proposed \height~representation is essential to the hard shadow rendering in a 2D image, and can be useful in other applications as well.
In this section, we propose several methods to obtain the \height~map.

\noindent \textbf{Calculated from 3D geometries.} Given a 3D geometry and camera parameters, the \height~map can be computed by calculating the projection of the distance between each projected point and its footpoint. Fig.~\ref{Fig.real}-(b) illustrates the rendererd RGB image and its \height~map. With an accurate \height~map, the proposed approach can render realistic soft shadows in real-time and generate visually comparable results with the renderings from a physics-based renderer (see Fig.~\ref{Fig.benchmark}).

\noindent \textbf{Labeled from 2D images.} The \height~could also be annotated from a real RGB image by experienced annotators. Annotators are required to label sparse points on the object masks. For each point, its perpendicular footpoint on the ground is annotated. Thus, the \height~could be calculated by the distance along the y-axis. Bi-linear interpolation is employed to get the dense \height~map for the object of interest. Although the interpolation method is not physically correct, the generated hard shadows with the interpolated dense \height~maps are perceptually pleasing. Fig.~\ref{Fig.real}-(c) illustrates the sparsely labeled RGB images and interpolated \height~maps.  

\noindent \textbf{Estimated from 2D images.} Similar to the monocular depth estimation~\cite{saxena2005learning}, estimating the \height~from a single view image is an ill-posed problem. We verify that the \height~could be estimated from a single view image. We propose a neural network for estimating humans' \height. 
The input to the network is the concatenation of the foreground image, the object mask, and a Y-Coordinate Map (YCM). 
We normalize the YCM by setting the lowest point in the object mask to be zero. 
\height~map estimation is a high-level prediction problem, and the network should encode global information to get a better understanding of the geometry of the object. We employ an off-the-shelf transformer backbone, Mix Transformer encoder (MiT)~\cite{xie2021segformer}. A simple decoder merges features from different scales. The network's output is a one-channel \height~map. 
For the training, We minimize the mean square error for each pixel inside the object mask between the prediction and the ground truth \height~map. A total variation loss is used to regularize the prediction.

We use a synthetic dataset consisting of $60K$ renderings of 3D human models with various poses. We name this dataset  \textit{Synthetic60K}. 
To improve the model's generalization on real images, $1,500$ real images are collected and sparsely annotated to build a benchmark named \textit{Real1500}. We used $1,000$ ($500$) images as the training (validation set). The ground truth \height~of Synthetic60K and Real1500 are generated based on the methods described earlier in this section. We merge the \textit{Synthetic60K} and the  \textit{Real1500} training set to train a \height~Estimation Network (HENet). 
Each mini-batch is evenly sampled from the the two datasets. 
More implementation details about the training are in the supplementary materials.

\subsection{Soft Shadow Generator}
\label{sec:soft_render}
With the \height~map and the proposed hard shadow renderer, a hard shadow map can be generated given a point light position in the image. %
To add softness to the shadow, we train a soft shadow generator to create the effect of an area light and control the softness based on user input. 

\begin{figure}[t]
\setlength{\belowcaptionskip}{-0.2cm}
\centering  %
\includegraphics[width=\linewidth]{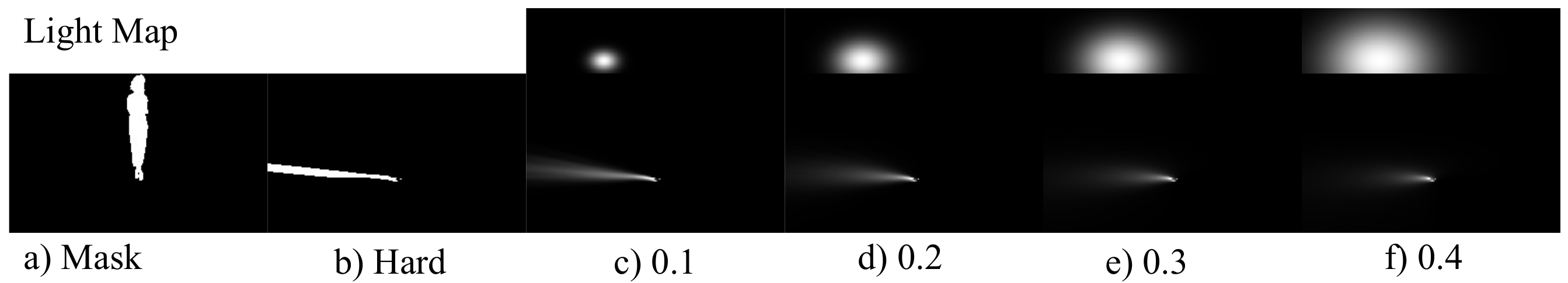}
\caption{The training set for the soft shadow generator.  a) the mask of the object. b) the hard shadow of the object. Figures c)-f) show the soft shadows and visualized light maps at different levels of softness. Softness models the size of the light source.} 
\label{Fig.softdata}
\end{figure}

\noindent\textbf{Data generation.} The soft shadows are generated following the pipeline in the Soft Shadow Network (SSN)~\cite{sheng2021ssn}. They divided the location of the light into grids, and then randomly sampled an environment light map based on a 2D Gaussian distribution at one random grid. Soft shadow bases are generated by merging the hard shadows of a local patch for each grid. The hard shadow is generated by a GPU-based render with 3D models. Soft shadow based on the environment lightmap will be the weighted sum for the shadow bases. SSN enforces the network to learn a mapping from the lightmap and the soft shadow, which is very complicated. Different from their method, we want to render a soft shadow based on the hard shadow and a pre-define softness. To get our training samples, for each soft shadow and its paired environment lightmap, we find the corresponding hard shadow and the softness from the lightmap. The hard shadow is rendered with a given point light, which locates at the center of the area light. The softness is defined as the size of the Gaussian which is used to generate the lightmap. Fig.~\ref{Fig.softdata} illustrates that an environment lightmap can be represented as a hard shadow and a softness value. Finally, we get our training triplet (hard shadow, soft shadow, and softness) on the fly during training.

\noindent\textbf{Network structure.} The soft shadow generator (SSG) is a variant of the U-Net. Similar to the shadow render in SSN~\cite{sheng2021ssn}, the encoder of the network is composed of a series of $3 \times 3$ convolution layers. Skip connections are employed to capture the low-level features. SSN~\cite{sheng2021ssn} estimates the soft shadow based on the object mask and the environment lightmap. It requires the network to learn a complex mapping between the object shadow and the light source. In contrast, we use a physical model to render the hard shadow in 2D space (described in Sec.~\ref{sec:hard_render}). The input of the encoder network is the concatenation of the mask and the hard shadow. To inject a softness control into the network, we uniformly discretize the continuous softness into multiple bins in the log space and then sampled a soft Gaussian distribution on these bins following~\cite{cao2017estimating,yin2019enforcing}. Thus, a softness value can be represented by an embedding with a fixed dimension. Following~\cite{karras2019style}, the adaptive instance normalization is then employed in the decoder to take the softness embedding for the softness control. The training details follow~\cite{sheng2021ssn}.

%% file: paragraph/experiment.tex
Our system consists of several key components: the Height Estimation Network (HENet), the Hard Shadow Renderer (HSR), and the Soft Shadow Generator (SSG). We first validate the effectiveness of HENet on human images and then the soft shadow quality from SSG. Finally, a user study and qualitative comparisons are conducted to evaluate our full system on real images. 

HENet is trained to predict the \height~for human images in our current implementation. In the following experiments, unless otherwise specified, the \height~ maps for humans are predicted by our HENet. The \height~ maps for other general objects are manually labeled. The average labeling time for one object is about two minutes.

\begin{table}[t]
\centering
\caption{Effectiveness of each components in predicting the~\height. YCM: using normalized Y-coordinate Map as input. Real: training on the real and synthetic data. $\ell_{tv}$: training with the total variation loss. The metrics are evaluated on the sparse points labelled by annotators on natural images. Base: Employing Y-Coordinate Map as the Pixel Height.}

\resizebox{0.8\textwidth}{!}{
\small
\begin{tabular}{p{0.1\linewidth}|p{0.15\linewidth}p{0.15\linewidth}p{0.15\linewidth}|p{0.1\linewidth}p{0.1\linewidth}}
\hline
 & YCM      & Real  & $\ell_{tv}$ & Abs $\downarrow$ & rel $\downarrow$\\\hline 
 Base&&&&10.84&3.64\\ \hline
 
a      & \checkmark & \checkmark &            & 6.12  &2.01     \\ 
b      &            & \checkmark & \checkmark & 6.04   &1.98     \\ 
c      & \checkmark &            & \checkmark & 7.05  &2.34     \\ 
d      & \checkmark & \checkmark & \checkmark & 5.92    &1.94     \\ \hline
\end{tabular}}

\label{tab:height_pred}
\end{table}

\begin{figure}[t]
\centering  %
\includegraphics[width=0.85\textwidth]{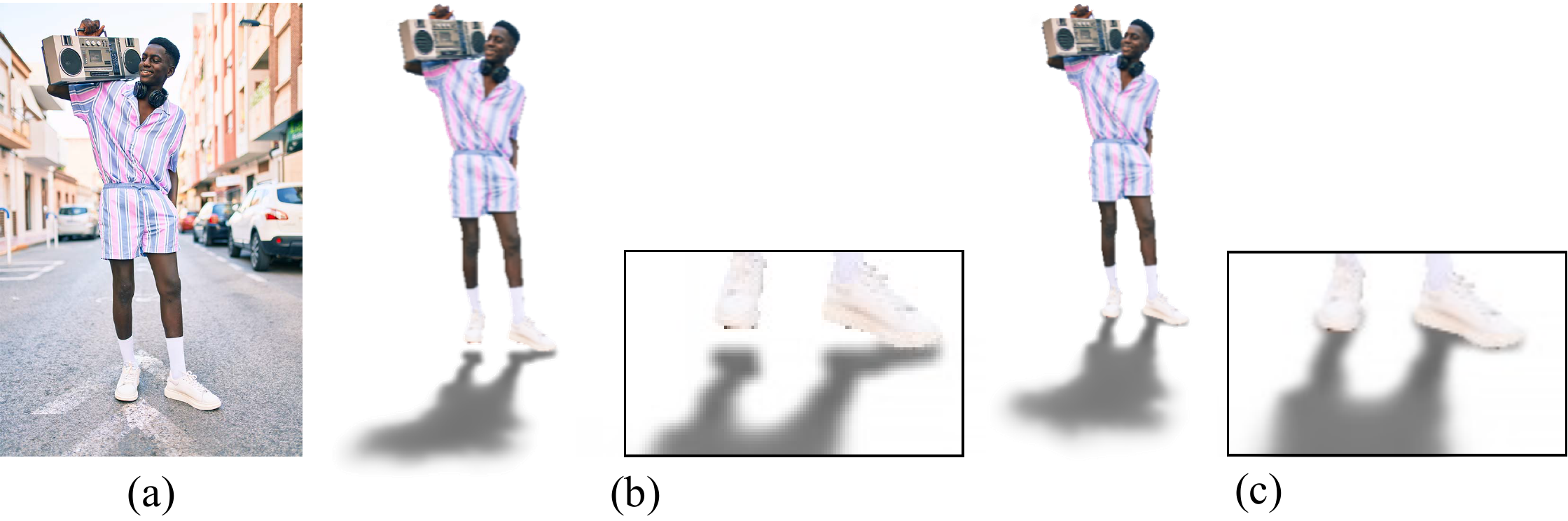}
\caption{For the input image without shadow (a) we use the Y-Coordinate Map to replace the \height~map in our system (b). It can not handle the foot contact with the ground properly. Based on our predicted \height~map, the shadow in the foot contact area is more realistic (c).}
\label{fig.fake}
\end{figure}

\subsection{Evaluation of HENet}
We conduct ablation studies on the proposed components to estimate the \height~map. The network is trained on the merged dataset of Synthetic60K and the Real1500 training set. The results are evaluated on the sparse points labeled by annotators of the 
Real1500 validation set. In Tab.~\ref{tab:height_pred}, the evaluation results show that adding the Y-Coordinate Map (YCM) and using the total variation loss ($\ell_{tv}$) can both reduce the error. Moreover, training on the merged dataset can significantly improve the model's generalization ability, reducing the relative error from $2.34\%$ to $1.82\%$. We also list the evaluation results of the baseline to verify that HENet does not just learn a trivial identity mapping of the YCM. As shown in Fig.~\ref{fig.fake}, using the YCM instead of the \height~map can not generate the correct shadow in the foot contact area.

\begin{figure}[t]
\centering  %
\includegraphics[width=0.95\linewidth]{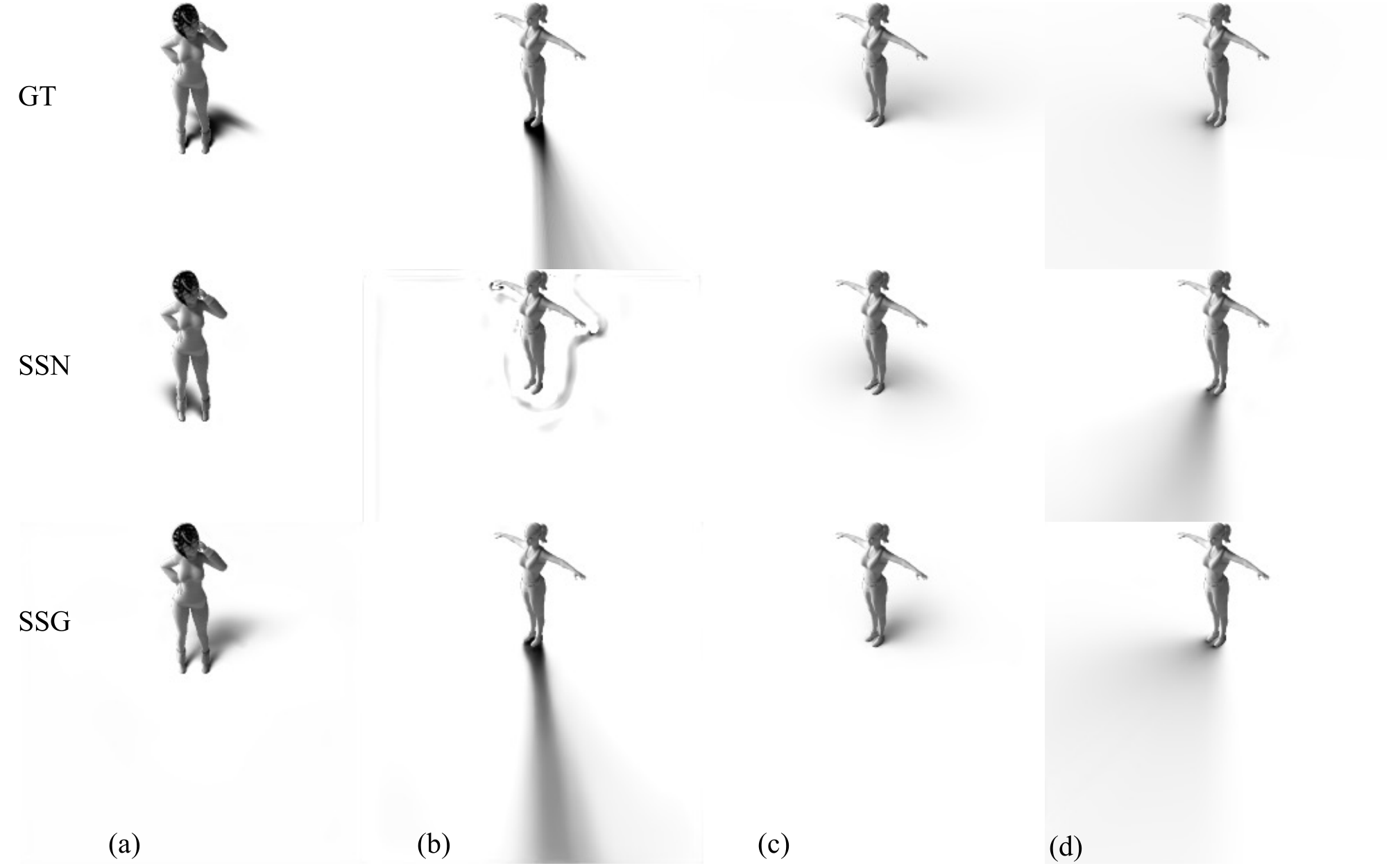}
\caption{Comparison between our proposed SSG, SSN~\cite{sheng2021ssn} and synthetic ground truth based on 3D models. (a) The direction of the shadow is more accurate as a hard shadow is given. (b) SSN may fail to generate a long hard shadow.  (c-d) Both methods perform well on soft shadows. The shadows from SSG are comparable to the physics-based renderer.}  
\label{Fig.benchmark}
\vspace{-2em}
\end{figure}

\subsection{Evaluation of SSG} 

\begin{table}[t]
\caption{\textbf{Quantitative evaluation.} Abs: pixel-level absolute error. ZNCC: zero normalized cross-correlation. The ground truth shadows are rendered with Mitsuba given 3D shapes. On average, our proposed SSG outperforms SSN~\cite{sheng2021ssn}. The quality of the `Long' shadow and the `Hard' shadow is improved with a larger margin.} 
\centering
\setlength{\tabcolsep}{5pt}
\footnotesize

\resizebox{0.75\textwidth}{!}{
\begin{tabular}{l|ccc|ccc}
\hline
         & \multicolumn{3}{c|}{Mean Abs} & \multicolumn{3}{c}{Mean ZNCC} \\ \hline
SSN~\cite{sheng2021ssn}      & \multicolumn{3}{c|}{0.033}         & \multicolumn{3}{c}{0.370}          \\
Ours      & \multicolumn{3}{c|}{\textbf{0.024}}         & \multicolumn{3}{c}{\textbf{0.788}}          \\ \hline  
\hline
         & \multicolumn{3}{c|}{Abs $\downarrow$}      & \multicolumn{3}{c}{ZNCC $\uparrow$}      \\\hline
Length   & Long     & Medium    & Short  & Long     & Medium    & Short  \\ \hline
SSN~\cite{sheng2021ssn}     & 0.041    & 0.029     & \textbf{0.031}  & 0.330    & 0.311     & 0.437  \\
Ours      & \textbf{0.028}    & \textbf{0.012}     & 0.033  & \textbf{0.743}    & \textbf{0.883 }    & \textbf{0.725}  \\ \hline \hline
Softness & Hard     & Medium    & Soft   & Hard     & Medium    & Soft   \\ \hline
SSN~\cite{sheng2021ssn}      & 0.039  & 0.034     & 0.024 & 0.198    & 0.336     & 0.606  \\
Ours      & \textbf{0.025 }   & \textbf{0.028 }    & \textbf{0.017}  & \textbf{0.761}    & \textbf{0.779 }    & \textbf{0.834}  \\ \hline
\end{tabular}
}
\label{tab:syn}
\end{table}

Instead of implicitly learning a mapping from the light source to a shadow~\cite{sheng2021ssn}, our SSG only focuses on adding softness to the hard shadow based on a controllable input scalar.
We build an evaluation benchmark to evaluate the model. We used $20$ new assets of 3D models that have no overlap with the training set, and they are collected from the Internet. For each new asset, we uniformly sample $4\times{6}6$ positions of the light source and divide them into three groups on average based on the length of the generated shadows, named as `Short', `Medium', and `Long'. For each position, $9$ types of softness are sampled. The evaluation benchmark is also divided into `Soft', `Medium', and `Hard' based on the softness. The ground truth shadows are rendered with Mitsuba with 3D models. The evaluation metrics include the average of the pixel-level absolute error (Abs) and the zero normalized cross-correlation (ZNCC). The first one evaluates the pixel-level error, and the second one considers the similarity of the shape.

\noindent\textbf{Results.} The evaluation results are shown in Tab.~\ref{tab:syn}. On average, the proposed SSG outperforms SSN on both evaluation metrics, improving Abs and ZNCC by $27\%$ and $112\%$, respectively. For hard shadows, the proposed SSG reduces the Abs error of SSN from $0.039$ to $0.025$, and increases the ZNCC from $0.198$ to $0.761$. The SSN performs slightly better on `Short' shadows according to the Abs, indicating that directly learning the mapping has some advantage in those cases. Still, it gets unstable for generating long and hard shadows due to the lack of model capacity in long-distance geometric modeling. 
Samples of visualization results are shown in Fig.~\ref{Fig.benchmark}. SSN may produce inaccurate direction of the shadow based on the given lightmap as shown in Fig.~\ref{Fig.benchmark}-(a). The errors on harder shadows are more apparent, and people are less sensitive to the difference in soft shadows.

\begin{table}[t]
\centering
\caption{\textbf{User study on natural images.} In a 2AFC study the users chose the more realistic image from a pair. The results indicate that 74\% of users perceived the shadows generated by our algorithm as more realistic.}
	\setlength{\tabcolsep}{4pt}
	\footnotesize

\resizebox{0.75\textwidth}{!}{
\begin{tabular}{c|ccc|ccc|c}
\hline
\multirow{2}{*}{Rate} & \multicolumn{3}{c|}{Length} & \multicolumn{3}{c|}{Softness}                                                     & \multicolumn{1}{c}{\multirow{2}{*}{Mean}} \\ \cline{2-7}
                      & Long   & Medium  & Short   & \multicolumn{1}{c}{Hard} & \multicolumn{1}{c}{Medium} & \multicolumn{1}{c|}{Soft} & \multicolumn{1}{c}{}                      \\ \cline{1-8} 
SSN                   & 0.27       &0.22          &0.35         &    0.20                      &                0.29            &     0.31                      &  0.26                                         \\
SSG                   & \textbf{0.73}       &\textbf{0.78 }         &\textbf{0.65}         &\textbf{0.80 }                         &\textbf{0.71 }                           &      \textbf{ 0.69  }                  &\textbf{0.74  }                                         \\ \hline
\end{tabular}
}

\label{tab:usr}
\vspace{-1em}
\end{table}

\subsection{Full System Evaluation}
We qualitatively evaluated our entire system on natural images. Specifically, we performed a user study, where we asked human subjects to compare the perceived visual quality of the generated shadows from our method and SNN. 

\begin{figure}[t]
\centering  %
\includegraphics[width=1\linewidth]{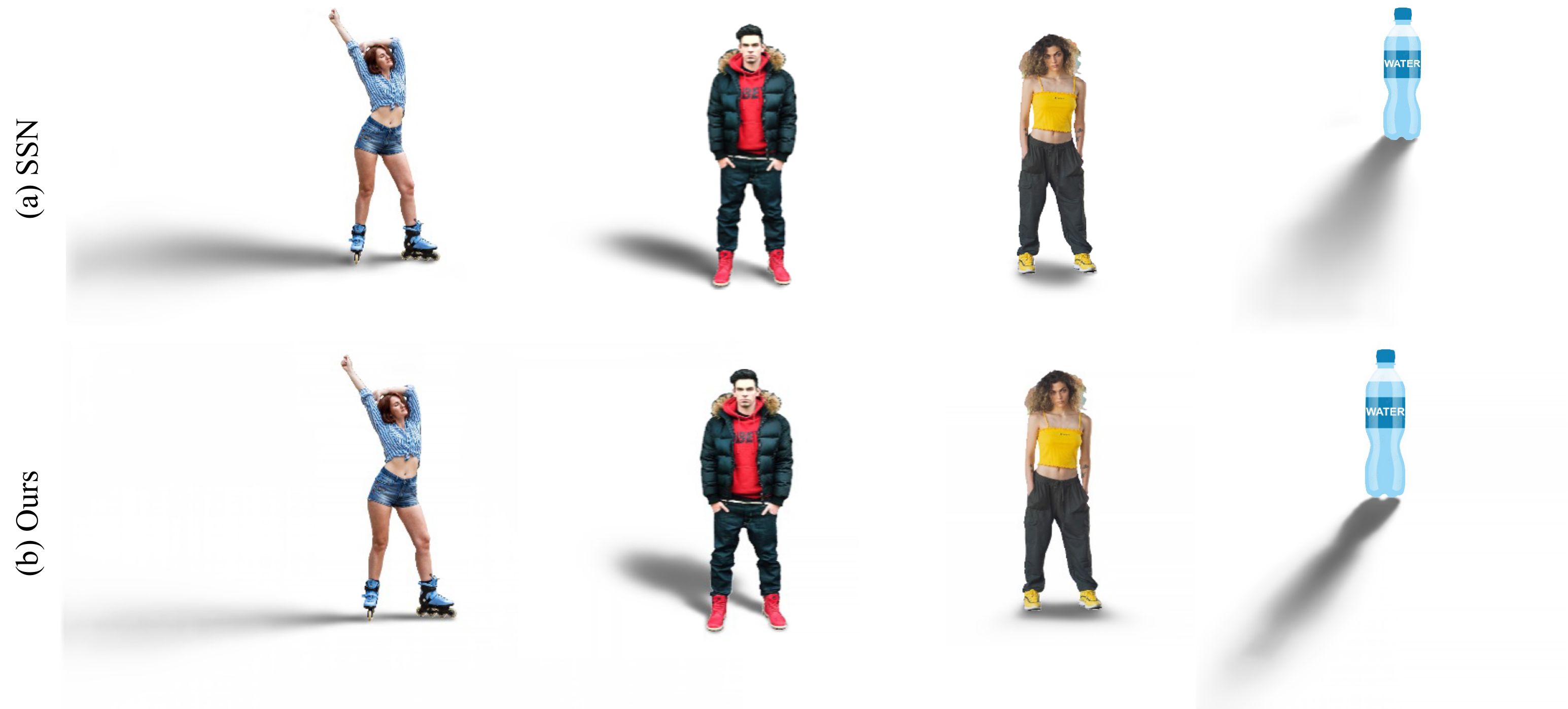}
\caption{\textbf{Samples images from our user study}. The shadows are generated on natural images in the wild with similar light. Our results have clearer shapes on hard shadows and less artifact on long shadows compared with SSN~\cite{sheng2021ssn}.}  
\label{Fig.user}
\end{figure}

We conducted a user study on the shadows generated for 2D natural images. For SSN, the shadow is rendered with the cutout of the object and an interactive light source. For our method, the shadow is rendered with an interactive interface with shadow position and softness controls (see the video demo in our supplementary materials). 
We prepared $24$ shadow pairs mimicking the effect of different lengths and softness. 
We have shown pairs of images in random order and random position (left-right) to $50$ users ($80\%$ males and $20\%$ females) and asked the participants which of the two images looks more realistic. Tab.~\ref{tab:usr} shows that $74\%$ of the users perceived the shadows rendered with our method as more realistic, especially for long and hard shadows (see Fig.~\ref{Fig.user}).

\begin{figure}[t]
\setlength{\belowcaptionskip}{-0.2cm}
\centering  %
\includegraphics[width=1\linewidth]{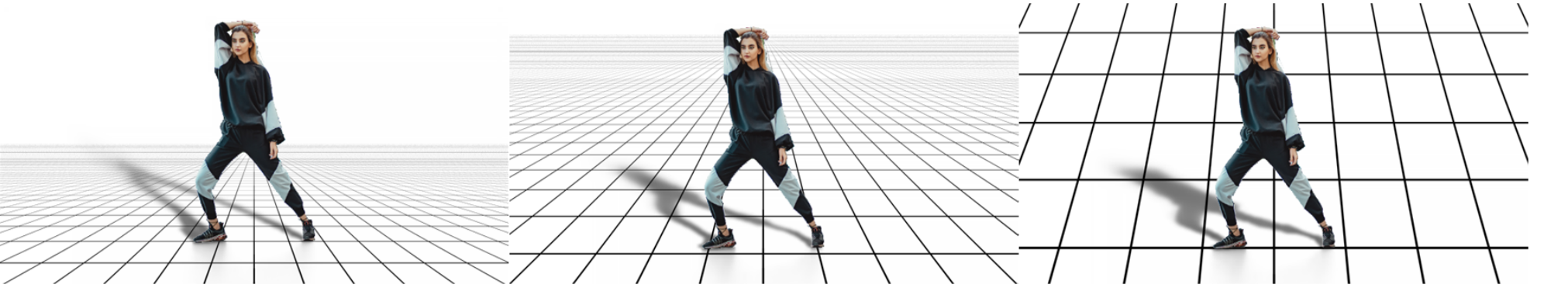}
\vspace{-1.5em}
\caption{\textbf{Controllability.} We can mimic the shadow effect for different camera poses by changing the horizon line.}
\label{Fig.hard}
\end{figure}

\section{Discussions}
\begin{figure}[t]
\centering  %
\includegraphics[width=0.9\linewidth]{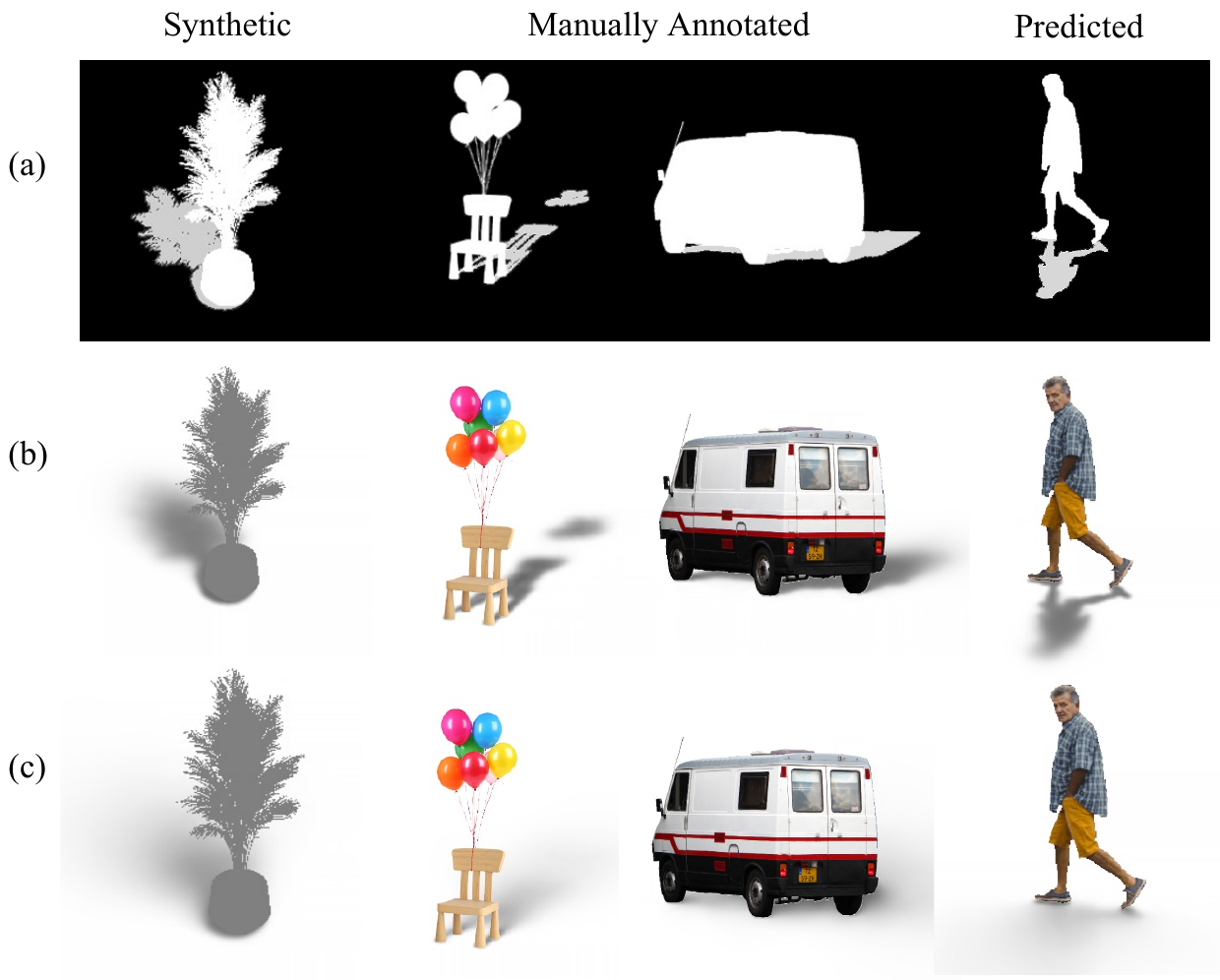}
\caption{Given a \height~map, our method can produce realistic shadows with desired position and softness. The \height~map can be calculated from 3D models, manually annotated or predicted by HENet. (a) Hard shadow mask. (b) Softness is 0.05. (c) Softness is 0.4.}
\label{Fig.differentpixel}
\end{figure}

\begin{figure}[hbt!]
\setlength{\tabcolsep}{0.0pt}
\begin{tabular}{cccc}
\includegraphics[width=0.249\linewidth]{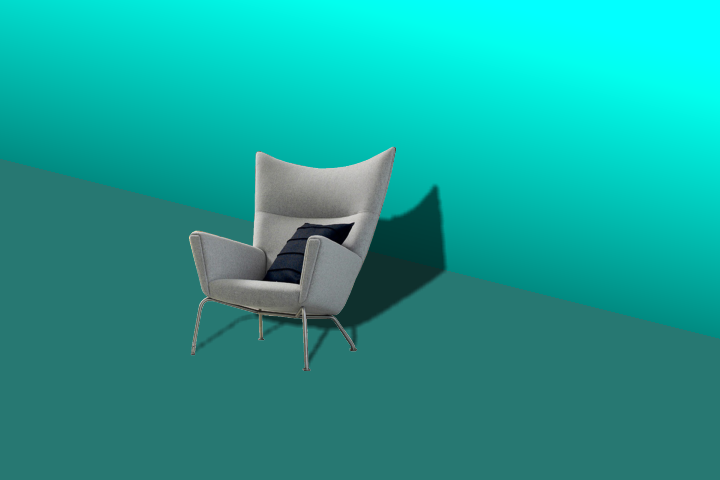} &
\includegraphics[width=0.249\linewidth]{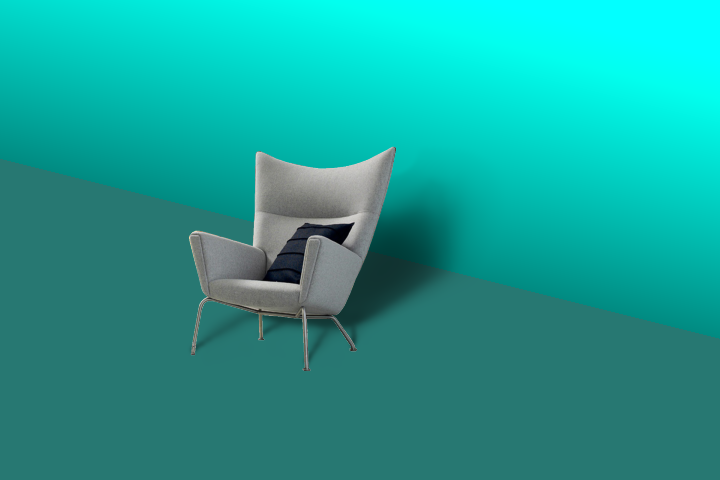} &
\includegraphics[width=0.249\linewidth]{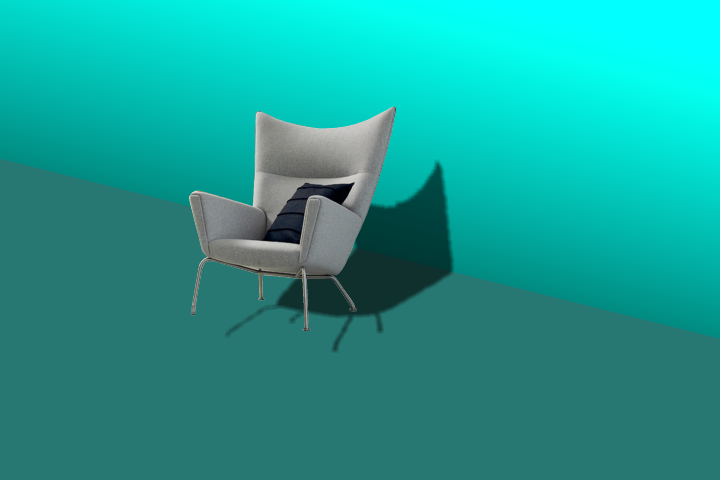} &
\includegraphics[width=0.249\linewidth]{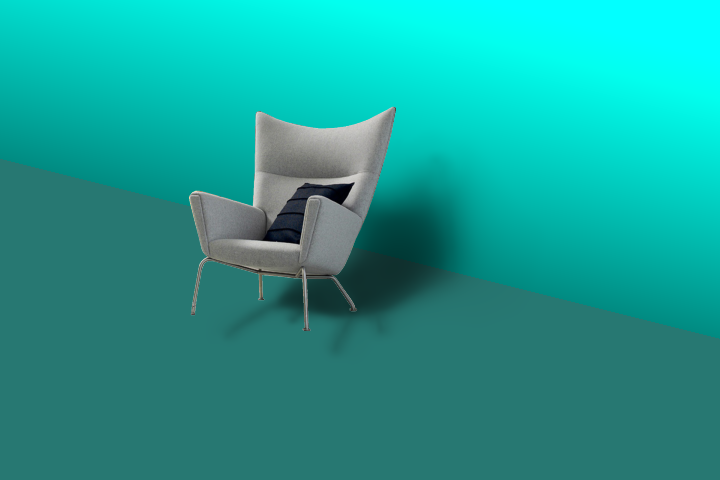} \\
(a) & (b) & (c) & (d)
\end{tabular}
\caption{Shadow generation for a floating object and a complex shadow receiver. (a) and (c) show that hard shadows can be cast on a non-planar shadow receiver using the \height~map of background as input. Our hard shadow renderer can also render shadows for floating objects by simply adding a shift value to the \height~map of the object. (b) and (d) show the corresponding soft shadow generated by SSG.
}  
\label{Fig.shadow_wall}
\end{figure}

\noindent\textbf{Controllability.} Our system based on \height~improves the controllability of the shadow synthesis. 
A demo video of our simple GUI is in the \textsl{supplementary materials}, enabling users to change the direction of the shadow by a simple click on the preferred position, similar to the method presented in \cite{pellacini2002user}. Our method also allows the control of the shadow shape using the horizon line, mimicking the perspective effect from a camera (see Fig.~\ref{Fig.hard}). The softness is controlled by a slider. 
Fig.~\ref{Fig.differentpixel} shows some example results generated from our GUI using height maps obtained from different approaches. Fig.~\ref{Fig.shadow_wall} shows a case where the object's shadow is cast on a complex shadow receiver with a floating effect. Our method can also be applied on animated objects. Please check out our supplementary materials for more examples.

\noindent\textbf{Potential applications of the \height~map.} \newcontent{\height~map can also be used to generate reflection effects. A slightly modified checking condition is used to compute the correspondence between a point and its reflection on the ground. We demonstrate this in the supplementary material.}

%% file: main.bbl
\begin{thebibliography}{10}
\providecommand{\url}[1]{\texttt{#1}}
\providecommand{\urlprefix}{URL }
\providecommand{\doi}[1]{https://doi.org/#1}

\bibitem{annen2008real}
Annen, T., Dong, Z., Mertens, T., Bekaert, P., Seidel, H.P., Kautz, J.:
  Real-time, all-frequency shadows in dynamic scenes. ACM TOG  \textbf{27}(3),
  ~1--8 (2008)

\bibitem{assarsson2003geometry}
Assarsson, U., Akenine-M{\"o}ller, T.: A geometry-based soft shadow volume
  algorithm using graphics hardware. ACM TOG  \textbf{22}(3),  511--520 (2003)

\bibitem{cao2017estimating}
Cao, Y., Wu, Z., Shen, C.: Estimating depth from monocular images as
  classification using deep fully convolutional residual networks. IEEE TCSVT
  \textbf{28}(11),  3174--3182 (2017)

\bibitem{chan2003rendering}
Chan, E., Durand, F.: Rendering fake soft shadows with smoothies. In: Rendering
  Techniques. pp. 208--218. Citeseer (2003)

\bibitem{cook1984distributed}
Cook, R.L., Porter, T., Carpenter, L.: Distributed ray tracing. In: ACM
  SIGGRAPH. pp. 137--145 (1984)

\bibitem{crow1977shadow}
Crow, F.C.: Shadow algorithms for computer graphics. ACM SIGGRAPH
  \textbf{11}(2),  242--248 (1977)

\bibitem{donnelly2006variance}
Donnelly, W., Lauritzen, A.: Variance shadow maps. In: Proceedings of the 2006
  symposium on Interactive 3D graphics and games. pp. 161--165 (2006)

\bibitem{fernando2005percentage}
Fernando, R.: Percentage-closer soft shadows. In: ACM SIGGRAPH, pp. 35--es
  (2005)

\bibitem{franke2014delta}
Franke, T.A.: Delta voxel cone tracing. In: 2014 IEEE International Symposium
  on Mixed and Augmented Reality (ISMAR). pp. 39--44. IEEE (2014)

\bibitem{fuchs1985fast}
Fuchs, H., Goldfeather, J., Hultquist, J.P., Spach, S., Austin, J.D.,
  Brooks~Jr, F.P., Eyles, J.G., Poulton, J.: Fast spheres, shadows, textures,
  transparencies, and imgage enhancements in pixel-planes. ACM SIGGRAPH
  \textbf{19}(3),  111--120 (1985)

\bibitem{guennebaud2006real}
Guennebaud, G., Barthe, L., Paulin, M.: Real-time soft shadow mapping by
  backprojection. In: Rendering techniques. pp. 227--234 (2006)

\bibitem{guennebaud2007high}
Guennebaud, G., Barthe, L., Paulin, M.: High-quality adaptive soft shadow
  mapping. In: Computer Graphics Forum. vol.~26, pp. 525--533. Wiley Online
  Library (2007)

\bibitem{heitz2016real}
Heitz, E., Dupuy, J., Hill, S., Neubelt, D.: Real-time polygonal-light shading
  with linearly transformed cosines. ACM TOG  \textbf{35}(4), ~1--8 (2016)

\bibitem{hong2021shadow}
Hong, Y., Niu, L., Zhang, J., Zhang, L.: Shadow generation for composite image
  in real-world scenes. arXiv preprint arXiv:2104.10338  (2021)

\bibitem{hu2019mask}
Hu, X., Jiang, Y., Fu, C.W., Heng, P.A.: Mask-shadowgan: Learning to remove
  shadows from unpaired data. In: ICCV. pp. 2472--2481 (2019)

\bibitem{isola2017image}
Isola, P., Zhu, J.Y., Zhou, T., Efros, A.A.: Image-to-image translation with
  conditional adversarial networks. In: CVPR. pp. 1125--1134 (2017)

\bibitem{jiang2021ssh}
Jiang, Y., Zhang, H., Zhang, J., Wang, Y., Lin, Z., Sunkavalli, K., Chen, S.,
  Amirghodsi, S., Kong, S., Wang, Z.: Ssh: A self-supervised framework for
  image harmonization. In: ICCV. pp. 4832--4841 (2021)

\bibitem{kajiya1986rendering}
Kajiya, J.T.: The rendering equation. In: ACM SIGGRAPH. pp. 143--150 (1986)

\bibitem{karras2019style}
Karras, T., Laine, S., Aila, T.: A style-based generator architecture for
  generative adversarial networks. In: CVPR. pp. 4401--4410 (2019)

\bibitem{li2018megadepth}
Li, Z., Snavely, N.: Megadepth: Learning single-view depth prediction from
  internet photos. In: CVPR. pp. 2041--2050 (2018)

\bibitem{liu2020arshadowgan}
Liu, D., Long, C., Zhang, H., Yu, H., Dong, X., Xiao, C.: Arshadowgan: Shadow
  generative adversarial network for augmented reality in single light scenes.
  In: CVPR. pp. 8139--8148 (2020)

\bibitem{liu2018auto}
Liu, Y., Qin, Z., Wan, T., Luo, Z.: Auto-painter: Cartoon image generation from
  sketch by using conditional wasserstein generative adversarial networks.
  Neurocomputing  \textbf{311},  78--87 (2018)

\bibitem{lu2019indices}
Lu, H., Dai, Y., Shen, C., Xu, S.: Indices matter: Learning to index for deep
  image matting. In: ICCV. pp. 3266--3275 (2019)

\bibitem{mehta2012axis}
Mehta, S.U., Wang, B., Ramamoorthi, R.: Axis-aligned filtering for interactive
  sampled soft shadows. ACM TOG  \textbf{31}(6),  1--10 (2012)

\bibitem{ng2003all}
Ng, R., Ramamoorthi, R., Hanrahan, P.: All-frequency shadows using non-linear
  wavelet lighting approximation. In: ACM SIGGRAPH, pp. 376--381 (2003)

\bibitem{oztireli2016integration}
{\"O}ztireli, A.C.: Integration with stochastic point processes. ACM TOG
  \textbf{35}(5),  1--16 (2016)

\bibitem{pellacini2002user}
Pellacini, F., Tole, P., Greenberg, D.P.: A user interface for interactive
  cinematic shadow design. ACM TOG  \textbf{21}(3),  563--566 (2002)

\bibitem{reeves1987rendering}
Reeves, W.T., Salesin, D.H., Cook, R.L.: Rendering antialiased shadows with
  depth maps. In: ACM SIGGRAPH. pp. 283--291 (1987)

\bibitem{remondino2003point}
Remondino, F.: From point cloud to surface: the modeling and visualization
  problem. International Archives of the Photogrammetry, Remote Sensing and
  Spatial Information Sciences  \textbf{34} (2003)

\bibitem{ren2006real}
Ren, Z., Wang, R., Snyder, J., Zhou, K., Liu, X., Sun, B., Sloan, P.P., Bao,
  H., Peng, Q., Guo, B.: Real-time soft shadows in dynamic scenes using
  spherical harmonic exponentiation. In: ACM SIGGRAPH, pp. 977--986 (2006)

\bibitem{pifuSHNMKL19}
Saito, S., , Huang, Z., Natsume, R., Morishima, S., Kanazawa, A., Li, H.: Pifu:
  Pixel-aligned implicit function for high-resolution clothed human
  digitization. ICCV  (2019)

\bibitem{saito2020pifuhd}
Saito, S., Simon, T., Saragih, J., Joo, H.: Pifuhd: Multi-level pixel-aligned
  implicit function for high-resolution 3d human digitization. In: CVPR (June
  2020)

\bibitem{saxena2005learning}
Saxena, A., Chung, S.H., Ng, A.Y., et~al.: Learning depth from single monocular
  images. In: NeurIPS. vol.~18, pp.~1--8 (2005)

\bibitem{schwarz2007bitmask}
Schwarz, M., Stamminger, M.: Bitmask soft shadows. In: Computer Graphics Forum.
  vol.~26, pp. 515--524. Wiley Online Library (2007)

\bibitem{sen2003shadow}
Sen, P., Cammarano, M., Hanrahan, P.: Shadow silhouette maps. ACM TOG
  \textbf{22}(3),  521--526 (2003)

\bibitem{sheng2021ssn}
Sheng, Y., Zhang, J., Benes, B.: {SSN}: Soft shadow network for image
  compositing. In: CVPR. pp. 4380--4390 (2021)

\bibitem{sillion1991global}
Sillion, F.X., Arvo, J.R., Westin, S.H., Greenberg, D.P.: A global illumination
  solution for general reflectance distributions. In: ACM SIGGRAPH. pp.
  187--196 (1991)

\bibitem{soler1998fast}
Soler, C., Sillion, F.X.: Fast calculation of soft shadow textures using
  convolution. In: ACM SIGGRAPH. pp. 321--332 (1998)

\bibitem{westin1992predicting}
Westin, S.H., Arvo, J.R., Torrance, K.E.: Predicting reflectance functions from
  complex surfaces. In: ACM SIGGRAPH. pp. 255--264 (1992)

\bibitem{williams1978casting}
Williams, L.: Casting curved shadows on curved surfaces. In: ACM SIGGRAPH. pp.
  270--274 (1978)

\bibitem{xie2021segformer}
Xie, E., Wang, W., Yu, Z., Anandkumar, A., Alvarez, J.M., Luo, P.: Segformer:
  Simple and efficient design for semantic segmentation with transformers.
  NeurIPS  (2021)

\bibitem{yin2021virtual}
Yin, W., Liu, Y., Shen, C.: Virtual normal: Enforcing geometric constraints for
  accurate and robust depth prediction. IEEE TPAMI  (2021)

\bibitem{yin2019enforcing}
Yin, W., Liu, Y., Shen, C., Yan, Y.: Enforcing geometric constraints of virtual
  normal for depth prediction. In: ICCV. pp. 5684--5693 (2019)

\bibitem{zhang2019shadowgan}
Zhang, S., Liang, R., Wang, M.: Shadowgan: Shadow synthesis for virtual objects
  with conditional adversarial networks. Computational Visual Media
  \textbf{5}(1), ~8 (2019)

\end{thebibliography}
